\documentclass[journal]{IEEEtran}
\usepackage{amsmath,amsfonts}
\usepackage{algorithmic}
\usepackage{algorithm}
\usepackage{amsmath}
\usepackage{amssymb}
\usepackage{graphicx}
\usepackage{array}
\usepackage[caption=false,font=normalsize,labelfont=sf,textfont=sf]{subfig}
\usepackage{textcomp}
\usepackage{stfloats}
\usepackage{url}
\usepackage{verbatim}
\usepackage{graphicx}
\usepackage{cite}
\usepackage{tabularx}
\usepackage{booktabs}
\usepackage{rerunfilecheck}
\usepackage[colorlinks=true,linkcolor=blue,citecolor=blue,urlcolor=blue]{hyperref}

\begin{document}

\title{RAZER: Robust Accelerated Zero-Shot 3D Open-Vocabulary Panoptic Reconstruction with Spatio-Temporal Aggregation}
\author{Naman Patel$^{1}$, Prashanth Krishnamurthy$^{1}$, Farshad Khorrami$^{1}$
    \thanks{$^{1}$Control/Robotics Research Laboratory (CRRL), Department of Electrical and Computer Engineering, NYU Tandon School of Engineering, Brooklyn, NY, 11201. E-mails: \{\texttt{nkp269, prashanth.krishnamurthy,khorrami}\}@nyu.edu}
}


\maketitle

\begin{abstract}
Mapping and understanding complex 3D environments is fundamental to how autonomous systems perceive and interact with the physical world, requiring both precise geometric reconstruction and rich semantic comprehension. While existing 3D semantic mapping systems excel at reconstructing and identifying predefined object instances, they lack the flexibility to efficiently build semantic maps with open-vocabulary during online operation. Although recent vision-language models have enabled open-vocabulary object recognition in 2D images, they haven't yet bridged the gap to 3D spatial understanding. The critical challenge lies in developing a training-free unified system that can simultaneously construct accurate 3D maps while maintaining semantic consistency and supporting natural language interactions in real time. In this paper, we develop a zero-shot framework that seamlessly integrates GPU-accelerated geometric reconstruction with open-vocabulary vision-language models through online instance-level semantic embedding fusion, guided by hierarchical object association with spatial indexing. Our training-free system achieves superior performance through incremental processing and unified geometric-semantic updates, while robustly handling 2D segmentation inconsistencies. The proposed general-purpose 3D scene understanding framework can be used for various tasks including zero-shot 3D instance retrieval, segmentation, and object detection to reason about previously unseen objects and interpret natural language queries. The project page is available at  \href{https://razer-3d.github.io/}{razer-3d.github.io}.
\end{abstract}


\section{Introduction}
\label{sec:intro}
\IEEEPARstart{T}{he} ability to create semantically meaningful 3D maps of dynamic environments is crucial for applications ranging from robotic navigation and manipulation to augmented reality and scene understanding. While significant advances have been made in geometric 3D reconstruction and 2D semantic understanding separately, combining these capabilities into a real-time system that can handle arbitrary objects and support natural language interactions remains a fundamental challenge in computer vision and robotics. This integration is essential for enabling intelligent systems to not only perceive the geometric structure of their environment but also understand and reason about the objects and their relationships within it.

Traditional 3D mapping systems have primarily focused on geometric accuracy, employing techniques like simultaneous localization and mapping (SLAM) and dense reconstruction to create precise spatial representations. However, these approaches typically lack semantic understanding, limiting their utility in applications requiring object-level reasoning. 3D semantic-instance mapping addresses this limitation by simultaneously detecting and segmenting objects with their semantic and instance labels to generate a comprehensive 3D semantic map of the surrounding environment. 
\begin{figure}[!t]
    \centering
    \includegraphics[width = \linewidth]{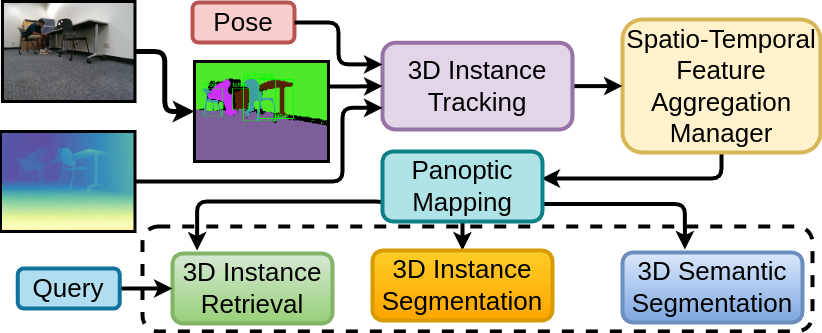}
    \caption{\textbf{Pipeline overview of our proposed 3D scene understanding framework}. Our system processes posed RGB-D inputs through open-vocabulary segmentation for robust 3D instance tracking. Spatio-temporal feature aggregation fuses and prunes tracks while updating a panoptic map that enables online text-based 3D instance retrieval and segmentation tasks.}
    \label{fig:overview}
\end{figure}
Existing approaches for 3D scene understanding broadly fall into two categories: \textit{3D-to-3D} and \textit{2D-to-3D} methods. The \textit{3D-to-3D} approach operates directly on dense 3D point clouds or volumetric representations, typically acquired through depth sensors or multi-view reconstruction, to perform object or concept level segmentation of the scene in 3D space. While these approaches benefit from having complete 3D geometric information available, they often struggle with computational efficiency and real-time performance due to the dense nature of point cloud processing as well as a scarcity of large-scale training data.  In contrast, \textit{2D-to-3D} methods analyze a set of 2D images and project their predictions onto the 3D map, performing reconstruction and semantic mapping synchronously. These approaches can leverage state-of-the-art 2D semantic-instance segmentation algorithms, partially addressing the limitations of \textit{3D-to-3D} methods. Recent works have attempted to bridge this gap by integrating semantic information into 3D reconstructions. However, these approaches face several critical challenges: they often require extensive training data, operate offline, or struggle with maintaining consistency when processing streaming data. Additionally, they frequently fail to handle the inherent inconsistencies in 2D segmentation outputs, leading to fragmented or incorrect 3D semantic maps. Furthermore, these approaches typically operate with predefined categories, limiting their applicability in open-world scenarios where systems must recognize and reason about previously unseen objects.

While recent breakthroughs in vision-language models have enabled remarkable open-vocabulary recognition capabilities in 2D images, these models lack the ability to reason about 3D structure and spatial relationships, creating a crucial gap in current advances. The fundamental challenges in creating a unified 3D semantic mapping system span multiple critical aspects of computer vision and robotics. A primary challenge lies in maintaining temporal consistency of object instances across frames without access to complete mapping history, as real-world applications often require processing streaming data with limited memory resources. This is compounded by the need to handle inconsistent labels and masks from 2D segmentation models in real-time, where prediction errors and uncertainties must be robustly managed. Additionally, enabling natural language interactions with the 3D environment without requiring task-specific training presents significant difficulties in bridging the gap between language understanding and spatial reasoning. These challenges are further complicated by the requirement to integrate geometric and semantic information in a computationally efficient manner suitable for real-time applications, where processing constraints demand careful optimization of both memory and computational resources.

\begin{figure*}[!t]
    \centering
    \includegraphics[width = \linewidth]{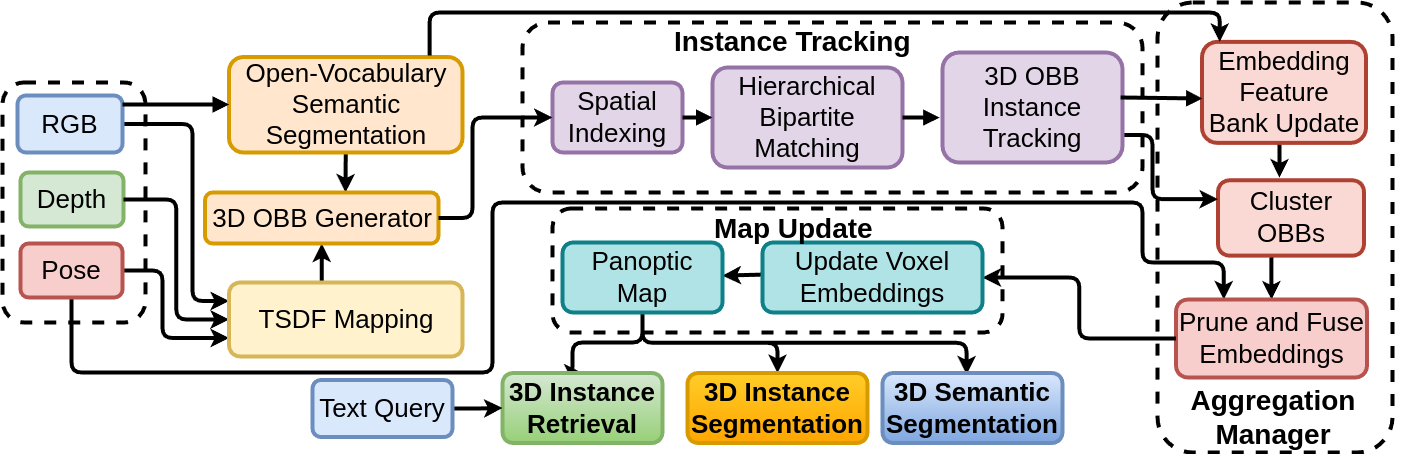}
    \caption{\textbf{System-level architecture of our RAZER framework}. It processes RGB, depth, and pose inputs through three modules: (1) \textbf{Instance Tracking} to enable efficient feature updates, (2) \textbf{Aggregation Manager} to aggregate and fuse/prune instances and their corresponding coarse features, and (3) \textbf{Map Update} to update features at voxel level and their corresponding labels, thus generating a panoptic map that enables 3D scene understanding.}
    \label{fig:system}
\end{figure*}

In this paper, we present a novel zero-shot framework that addresses these challenges by seamlessly integrating geometric reconstruction with open-vocabulary vision-language models. Our key insight is that by maintaining a unified semantic embedding space and employing efficient spatial indexing strategies, we can achieve robust real-time performance while handling the uncertainties inherent in 2D segmentation outputs. This approach enables natural language interaction with the 3D environment while maintaining both geometric and semantic consistency.
Our framework addresses these challenges through several key innovations that enable real-time, zero-shot 3D semantic mapping. The core of our approach lies in maintaining temporal consistency without requiring global optimization, achieved through an instance-level semantic embedding fusion combined with efficient spatial indexing and association strategies for fast 3D tracking. Our key technical contributions can be summarized as follows:
\begin{itemize}
\item A modular zero-shot 3D semantic mapping framework leveraging pretrained vision-language models to perform various open-vocabulary 3D scene understanding tasks without training or fine-tuning.
\item An online, instance-level geometric and semantic fusion algorithm for RGB-D streams enabling real-time mapping without global optimization.
\item A robust object association strategy combining R-tree spatial indexing with minimum-cost bipartite matching for fast 3D tracking to effectively handle inconsistent masks and labels from 2D segmentation.
\item A unified geometric-semantic update mechanism ensuring temporal consistency via instance-level tracking and supporting natural language interactions with the 3D environment.
\end{itemize}
We demonstrate the effectiveness of our approach through extensive experiments on multiple open-vocabulary 3D benchmarks, showing superior performance in tasks such as instance retrieval, semantic segmentation, and instance segmentation. We demonstrate that our framework successfully bridges the gap between geometric reconstruction and semantic understanding, enabling robust 3D scene understanding with natural language interaction capabilities in unconstrained environments. The ability to handle arbitrary objects without requiring task-specific training, combined with real-time processing of streaming data while maintaining both geometric and semantic consistency, makes our framework particularly suitable for real-world applications with dynamically changing scenes. The rest of the paper is organized as follows: Section~\ref{sec:related-works} covers related work, Section~\ref{sec:prob-form} presents the problem and Section~\ref{sec:methodology} describes our framework, Section~\ref{sec:experiments} shows experimental results, and Section~\ref{sec:conclusion} concludes the paper.

\section{Related Work}
\label{sec:related-works}
Traditional 3D scene understanding methods remain constrained by closed-set categories pre-defined during training~\cite{Wang19CVPR, Elich19GCPR, Lahoud19ICCV, Hou19CVPR, Yang19NIPS}, limiting their adaptability to new object classes. Recent research leverages vision-language models (VLMs) to attain \emph{open-vocabulary} recognition in 3D, expanding beyond fixed labels. For instance, OpenScene projects 3D points into CLIP space, enabling text-driven queries with zero-shot capability~\cite{WuLXYDY0ZT0GT24}, while Open3DIS and OpenMask3D aggregate multi-view 2D instance masks into coherent 3D segments and associate them with language embeddings~\cite{ZhuChen23,XuZWLCHB22}. These approaches achieve instance-level segmentation of novel classes but often rely on offline or batch processing, making them less suitable for real-time use. Open-vocabulary \emph{3D instance retrieval} has also been explored, notably by methods that fuse 2D text-aligned proposals with incremental 3D mapping~\cite{Maggio2024Clio}, yet many do not offer online performance for robotics.

Incremental mapping integrates 2D semantic cues (e.g., segmentation masks) into an evolving 3D representation as a robot explores~\cite{voxbloxplusplus,PanopticFusion}. Early systems like SemanticFusion~\cite{RunzBA18} demonstrated real-time 3D semantic mapping by fusing per-frame labels, while Voxblox++~\cite{voxbloxplusplus} and PanopticFusion~\cite{PanopticFusion} extended this to instance-level and panoptic segmentation. Although effective for known classes, these pipelines cannot accommodate unseen categories. Recent work has introduced post-hoc optimization or multi-view consistency~\cite{Han20CVPR,Chen21ICCV} to mitigate frame-level label noise, but still within a closed-set taxonomy. Our approach builds on the strengths of real-time fusion but adopts an open-vocabulary 2D model for segmentation, enabling online handling of novel objects without retraining. VLMs such as CLIP~\cite{RadfordKHRGASAM21} and BLIP~\cite{li2023blip} have accelerated progress in zero-shot classification and retrieval for 2D images. Their extension to 3D data includes methods like PointCLIP~\cite{zhang2022pointclip} and ULIP~\cite{xue2023ulip}, which unify point cloud and text embeddings. These methods can recognize categories not present in any 3D training set, but often operate offline and are computationally heavy. Meanwhile, approaches like ConceptGraphs~\cite{gu2024conceptgraphs} use multi-view images and 2D VLM predictions to label 3D points or clusters, yielding open-vocabulary scene representations. However, they typically rely on batch (offline) processing or expensive clustering, limiting real-time deployment. The recent INS-CONV~\cite{INS_CONV} combines the advantages of 2D-3D and 3D-3D procedures. While it builds the 3D model of the environment incrementally from RGB-D frames, it performs segmentation directly in the 3D space. This, however, comes at the cost of requiring 3D ground truth annotations for training.

Recent works have significantly improved the performance of open-vocabulary 3D segmentation using point clouds from LiDAR or depth data from RGB-D cameras~\cite{DingYXZBQ23,3dvlp,3dvista,chen2024vote2cap,ChenGNC21,ChenCN20,SchultEHLTL23,YueMSELSK24,zhou2024unid}. In addition, there have also been approaches fusing embeddings from open world VLMs like CLIP with neural radiance fields, Gaussian splats or implicit neural network based representations~\cite{kerr2023lerf,ZhangLA24,WangCL24,nguyen2024semantically} for generating a consistent segmentation of the scene.
The semantic information aids in both understanding the scene as well as improving the accuracy of the map and odometry for semantic SLAM~\cite{RosinolVAHCSGC21,XuLTBDL19,RunzBA18,NicholsonMS19,patel2018semantic,MorenoNSKD13,KongLTD23,PatelKKT19,HanLDY23,McCormacCBDL18}. End-to-end 3D methods that predict 3D semantic information directly from sensor modalities like camera and depth, or from intermediate representations like reconstructed point clouds or voxel-based representations, can capture complex 3D relationships. Recent efforts combine SLAM-based mapping with language models to produce richer 3D scene graphs for tasks like grounded query and high-level planning~\cite{SceneGraphFusion,werby2024hierarchical,Maggio2024Clio,gu2024conceptgraphs}. By coupling object-level mappings with language-based descriptions or relational knowledge, these systems enable queries such as \emph{``retrieve the red chair in the corner''} and facilitate complex reasoning. However, computational overhead and reliance on offline graph construction remain common hurdles.
Semantic mapping techniques, vital for scene graph generation, extend beyond 2D segmentation to encompass complex 3D structures. Reconstruction of 3D structures with object recognition and grouping remains a fundamental challenge in computer vision, especially in the context of 3D scene understanding. These methods help a large language model understand the scene through reasoning~\cite{wang2024llm,Shi_Dao_Cai_2024} for building grounded 3D large language models~\cite{MaYZ0LZH23,AzumaMKK22,AchlioptasAXEG20,grounded3dllm,3dllm,ll3da,YangC0MIFC24}.

The open vocabulary allows for greater flexibility in real-world applications and can help an LLM to understand the scene with multiple objects, which multi-modal LLMs hallucinate or fail to perceive~\cite{ChenMZXQYCF24}. However, these methods do not scale well in larger scenes, making them computationally intensive and memory-consuming, rendering them impractical for real-time segmentation in robotic applications. In parallel, large language models (LLMs) like GPT~\cite{achiam2023gpt} have shown potential for reasoning over text-based representations of a scene~\cite{koch2024open3dsg}, but bridging real-time 3D reconstruction with LLM-driven semantics remains under-explored.

Our work aims to unify these threads by performing online, incremental 3D semantic instance mapping \emph{without} closed-set constraints, using a pretrained vision-language model for open-vocabulary segmentation. In contrast to methods that require domain-specific 3D supervision or post-hoc processing, we achieve zero-shot instance detection and tracking on the fly. By maintaining object-level embeddings in a continuously updated 3D map, we further enable tasks such as real-time retrieval of novel objects, making it practical for robotic semantic mapping for navigation.

\section{Problem Formulation}
\label{sec:prob-form}
We present an open-vocabulary 3D scene understanding system that processes RGB-D streams with known camera poses to build semantically rich 3D maps. Unlike traditional systems constrained to a fixed taxonomy of object categories, our approach enables unrestricted object recognition and tracking through continuous semantic embedding spaces. At each timestep $t$, our system processes an RGB image $I_t^{rgb} \in \mathbb{R}^{H \times W \times 3}$, a depth image $I_t^{depth} \in \mathbb{R}^{H \times W}$, and a camera pose $P_t \in SE(3)$, where $H$ and $W$ denote the image height and width respectively.

Unlike traditional closed-set methods that rely on a predefined label space, our zero-shot approach harnesses a \emph{pretrained} vision-language segmentation model capable of recognizing a wide range of object categories. Specifically, rather than fine-tuning or distilling from this model, we directly leverage its zero-shot capability to produce semantic embeddings for any concept of interest. A key innovation lies in maintaining a unified representation that encodes both geometric and semantic properties in a continuous embedding space. Rather than performing discrete classification into predefined categories, we leverage vision-language models to embed objects in a high-dimensional semantic space $\mathcal{S} \in \mathbb{R}^d$, where semantically similar concepts are naturally clustered together, enabling new categories to be recognized at runtime without retraining. Our system therefore supports:

\begin{itemize}
    \item \textbf{Volumetric reconstruction} of the scene using a Truncated Signed Distance Function (TSDF) $\Phi: \mathbb{R}^3 \to \mathbb{R}$ with truncation distance $\tau$.
    \item \textbf{Object detection and 6-DoF tracking} with oriented bounding boxes (OBBs) $\mathcal{B} = \{b_i\}_{i=1}^N$, where $b_i = (c, R, s) \in \mathbb{R}^3 \times SO(3) \times \mathbb{R}^3$ denotes the center, orientation, and scale respectively.
    \item \textbf{Semantic embeddings} $\mathcal{E} = \{e_i\}_{i=1}^N$, where $e_i \in \mathbb{R}^d$ accumulates open-vocabulary features for object $i$ using a \emph{pretrained} vision-language model. 
    \item \textbf{Probabilistic object pruning} through confidence scores $\alpha_i \in [0,1]$, updated based on temporal consistency and semantic stability.
    \item \textbf{Online fusion of new observations} to incrementally refine both geometry and semantics, ensuring robust instance identity across multiple viewpoints.
\end{itemize}
As our system operates in a zero-shot manner, at each timestep, the system extracts semantic features from the incoming RGB-D frame using a frozen, pretrained 2D model, and fuses them with previously observed data in 3D space. This approach allows the model to generalize to novel object categories so long as they are recognizable by the pretrained 2D backbone. 

A fundamental challenge in this work is maintaining robust object tracking and coherent semantic labeling under such unconstrained open-vocabulary settings. When the system first detects an object, the initial embedding $e_i^0$ may reflect incomplete or occluded views. As more viewpoints become available, the system refines $e_i^t$ through a temporal fusion function $e_i^t = f\bigl(e_i^{t-1}, z_i^t\bigr)$, where $z_i^t$ denotes newly extracted semantic features at time $t$. This process ensures that object instance identity is preserved via geometric consistency and occlusion-aware updates, even when only partial observations are available in any single frame.

\section{Approach}
\label{sec:methodology}

\subsection{Volumetric Scene Reconstruction}
We discretize the environment into a volumetric grid and maintain a TSDF volume $V: \mathbb{R}^3 \to \mathbb{R}^6$. For each voxel at position $\mathbf{p}\in\mathbb{R}^3$, the stored tuple includes:
\begin{equation}
    V(\mathbf{p}) \;=\; \bigl\{\,d,\; w,\; \mathbf{c},\; l,\; h\,\bigr\},
\end{equation}
where $d \in [-\tau,\tau]$ is the truncated signed distance to the nearest surface, $w \in [0,1]$ is the accumulated confidence of $d$, $\mathbf{c}\in \mathbb{R}^3$ is an RGB color, $l \in \mathbb{N}$ is an instance label index, and $h$ is a histogram tracking how frequently each instance label has been observed in that voxel. 

At each incoming frame $(I_t^{rgb}, I_t^{depth})$, we fuse new depth measurements into the TSDF volume via standard weighted averaging:
\[
d_{\text{new}} \;=\; \omega_{\text{old}}\,d_{\text{old}} \;+\;\omega_{\text{new}}\,d_{\text{obs}},
\]
where $d_{\text{obs}}$ is the signed distance derived from the depth map, and $\omega_{\text{old}}, \omega_{\text{new}}$ are confidence weights. The voxel color $\mathbf{c}$ and instance histogram $h$ are also updated if the voxel lies within the truncated distance bound of a newly observed surface. This ensures that regions recognized as belonging to a particular object accumulate a consistent label history. 

Through this process, our system continuously refines a dense 3D representation that captures both geometry and instance labels. Unlike conventional methods that only update a fixed label, our open-vocabulary setting allows for dynamic incorporation of semantic cues from the pretrained model, even if new object classes appear over time.

\subsection{3D Object Detection}
Our detection and tracking system integrates both geometric and semantic cues by first processing each RGB frame with an open-vocabulary vision-language segmentation model $\mathcal{F}$. This model generates instance masks $M_t$ for \emph{any} objects it can visually separate, without being restricted to a finite set of categories:

\begin{equation}
    M_t = \mathcal{F}(I_t^{rgb}).
\end{equation}

Since this is a zero-shot approach, we do not perform any additional training or distillation; instead, we rely on the model's ability to segment objects of interest based on its pre-trained knowledge from vision-language data. In cluttered scenes, the model can produce a large number of masks, potentially including false positives. Hence, a confidence-based threshold or heuristic filters can be applied to remove implausible masks.

\paragraph{3D Point Extraction} For each mask $m_i$ in $M_t$, we extract the corresponding 3D points from the depth map $I_t^{depth}$. Specifically, for every pixel $(x,y)$ that lies within mask $m_i$, we back-project it into 3D space:
\begin{equation}
    \mathbf{P}_i = \{\pi^{-1}(x, y, I_t^{depth}(x,y)) \,\mid\, (x,y) \in m_i\},
\end{equation}
where $\pi^{-1}$ is the camera intrinsics-based inverse projection function. Accurate calibration is assumed for correct 3D point placement.

\paragraph{3D Clustering for Object Separation.} 
Although a single instance mask in 2D often corresponds to a single object, occlusions and overlapping masks can lead to mismatches between 2D and 3D object boundaries. To address this, we apply DBSCAN clustering in 3D on the points $\mathbf{P}_i$ to distinguish distinct object clusters based on spatial density. DBSCAN parameters (e.g., $\varepsilon$, the neighborhood radius, and the minimum point count) must be tuned to balance merging and splitting errors.

\paragraph{Oriented Bounding Box (OBB) Fitting.} 
For each resulting 3D cluster, we compute an oriented bounding box (OBB) using Principal Component Analysis (PCA). Given $\mathbf{P}_i$ for a cluster, we first compute the centroid $\mathbf{c}$:
\begin{equation}
    \mathbf{c} = \frac{1}{N} \sum_{j=1}^N \mathbf{P}_j,
\end{equation}
and the covariance matrix:
\begin{equation}
    \mathbf{C} = \frac{1}{N-1} \sum_{j=1}^N (\mathbf{P}_j - \mathbf{c})(\mathbf{P}_j - \mathbf{c})^T.
\end{equation}
Eigen-decomposition $\mathbf{C} = \mathbf{R}\mathbf{\Lambda}\mathbf{R}^T$ yields the principal axes $\mathbf{R} = [\mathbf{v}_1, \mathbf{v}_2, \mathbf{v}_3]$. We ensure a right-handed coordinate system by reorienting $\mathbf{v}_3$ if necessary via $\mathbf{v}_3 \leftarrow \mathbf{v}_1 \times \mathbf{v}_2$. We compute the OBB extents by projecting the points onto each principal axis $\mathbf{v}_i$:
\begin{equation}
    e_i = \max_{j}\bigl(\mathbf{v}_i^T \mathbf{P}_j\bigr) - \min_{j}\bigl(\mathbf{v}_i^T \mathbf{P}_j\bigr).
\end{equation}
Each object detection can thus be represented by the tuple $(\mathbf{c}, \mathbf{R}, \mathbf{s})$, describing the center, principal axes, and box dimensions respectively.

\subsection{R-Tree Based Hierarchical Association for Tracking}
\label{subsec:rtree}
Having obtained new OBB detections, we must associate them with tracked objects from previous frames. This must be efficient, as real-time systems can track dozens of objects simultaneously.

\paragraph{R-Tree Organization.}
We store bounding boxes of tracked objects in an R-tree, which enables spatial queries in expected $\mathcal{O}(\log n)$ time. Each tracked object's oriented bounding box $b_i = (c_i, R_i, s_i)$ is converted to an axis-aligned bounding box (AABB):
\begin{equation}
    \mathbf{b}_{\text{min}}, \mathbf{b}_{\text{max}} \;=\; \mathrm{AABB\_Enclose}\bigl(c_i, R_i, s_i\bigr),
\end{equation}
and this $(\mathbf{b}_{\text{min}}, \mathbf{b}_{\text{max}})$ is stored in a leaf node of the R-tree. Internal nodes recursively store the minimal AABBs enclosing their children, creating a spatial hierarchy to prune searches for overlapping or nearby objects.

\paragraph{Association via R-Tree Query.} 
For a new detection $b_j^{(\text{new})}$, we generate its AABB and query the R-tree to retrieve only those tracked objects whose AABBs intersect or lie within a small distance. This reduces the candidate set from the entire pool of tracked objects to a manageable subset.

\paragraph{Candidate Matching and Hungarian Algorithm.} 
Within this candidate subset, we resolve final matches using a bipartite matching approach:
\begin{equation}
M_{ij} \;=\; w_v \, V_{ij} \;+\; w_s \, S_{ij}, 
\quad
\min_{\mathbf{X}} \;\sum_{i,j} M_{ij}\,X_{ij}
\end{equation}
where
\begin{itemize}
    \item $V_{ij} = 1 - \mathrm{IoU}(\mathrm{OBB}_i,\,\mathrm{OBB}_j^{(\text{new})})$ is the geometric dissimilarity based on 3D IoU (intersection over union).
    \item $S_{ij} = \|\mathbf{e}_i - \mathbf{e}_j^{(\text{new})}\|_2$ represents a semantic distance measure. If the system has an embedding vector for each object, $S_{ij}$ could combine both shape and open-vocabulary features. See Section~\ref{subsec:emb-management} for more details.
    \item $X_{ij} \in \{0,1\}$ are the elements of assignment matrix $\mathbf{X}$ that impose a one-to-one matching constraint.
\end{itemize}
We solve for the assignment matrix $\mathbf{X}$ to match candidate detections with tracked object OBBs using the Hungarian (Kuhn-Munkres) algorithm~\cite{kuhn1955hungarian} in $\mathcal{O}(m^3)$ time, where $m$ is the number of candidates. Due to the R-tree query, $m$ is usually small, making real-time matching computationally feasible. Unmatched new detections spawn new tracks, while previously tracked objects that remain unmatched for $k$ consecutive frames are pruned unless their semantic confidence is high (to handle long occlusions). This pruning applies only within the current camera frustum; global object persistence is maintained through the semantic map, which accounts for long-term occlusions, viewpoint changes, and spatial displacement as the robot navigates across rooms.

\subsection{Incremental OBB Updates}
For each matched object, we refine its OBB incrementally by assimilating newly observed 3D points. This is particularly important when objects are only partially visible or change orientation over time.

\paragraph{Incremental Covariance Calculation.} 
Let an object $o_i$ at frame $t-1$ have a scatter matrix $\mathbf{S}_{t-1}$, centroid $\mathbf{c}_{t-1}$, and $N_{t-1}$ points accumulated thus far. A new detection in frame $t$ contributes $N_{\text{new}}$ points $\mathbf{P}_{\text{new}}$, with centroid $\mathbf{c}_{\text{new}}$ and scatter matrix $\mathbf{S}_{\text{new}}$. We update:

\begin{equation}
    N_{t} = N_{t-1} + N_{\text{new}},
\end{equation}
\begin{equation}
    \mathbf{c}_{t} = \frac{N_{t-1}\,\mathbf{c}_{t-1} + N_{\text{new}}\,\mathbf{c}_{\text{new}}}{N_{t}},
\end{equation}
\begin{equation}
    \mathbf{S}_{t} = \mathbf{S}_{t-1} 
    \;+\; \mathbf{S}_{\text{new}}
    \;+\; \frac{N_{t-1}\,N_{\text{new}}}{N_{t}}
    \bigl(\mathbf{c}_{t-1}-\mathbf{c}_{\text{new}}\bigr)
    \bigl(\mathbf{c}_{t-1}-\mathbf{c}_{\text{new}}\bigr)^T.
\end{equation}

The new covariance matrix is $\mathbf{C}_t = \mathbf{S}_t \,/\, (N_t - 1)$. By performing eigendecomposition on $\mathbf{C}_t$, we derive updated principal axes and extents, which can expand or contract the bounding box based on the newly visible parts of the object.

\paragraph{R-Tree Synchronization.}
Finally, once $\mathrm{OBB}_i$ is updated to $(\mathbf{c}_t, \mathbf{R}_t, \mathbf{e}_t)$, we convert it back to an AABB via
\[
\mathrm{AABB\_Enclose}\bigl(\mathbf{c}_t, \mathbf{R}_t, \mathbf{e}_t\bigr)
\]
and update $o_i$'s entry in the R-tree accordingly. This ensures that future queries accurately reflect the object's most current spatial extent. Since no additional supervision or distillation is used, this approach handles unseen object types in a zero-shot manner and remains flexible across changing scene conditions.

\subsection{Open-Vocabulary Semantic Embedding Management}
\label{subsec:emb-management}
At the core of our system is the management of open-vocabulary semantic embeddings for tracked objects. Each object $o_i$ maintains a semantic state consisting of an embedding bank $\mathcal{E}_i = \{\mathbf{e}_1, \mathbf{e}_2, \mathbf{e}_3\}$ containing up to three concept embeddings, along with corresponding confidence scores $\mathcal{C}_i = \{c_1, c_2, c_3\}$ where $c_j \in [0,1]$. This multi-embedding approach enables maintaining uncertain hypotheses about an object's identity, crucial for handling ambiguous cases and novel objects.

When an object is first detected at time $t$, we initialize its semantic representation by extracting vision-language features $\mathbf{f}_t \in \mathbb{R}^{H \times W \times d}$ from the backbone network. Given the instance mask $m_t$, we compute the initial embedding through mask-guided feature aggregation:

\begin{equation}
   \mathbf{e}_t = \frac{1}{|m_t|} \sum_{(x,y) \in m_t} \mathbf{f}_t(x,y)
\end{equation}

where $|m_t|$ denotes the number of valid pixels in the mask. This normalized pooling operation produces a fixed-dimensional embedding $\mathbf{e}_t \in \mathbb{R}^d$ that captures the object's semantic properties while being invariant to mask size.

As we accumulate observations, the embedding bank is updated according to semantic similarity and confidence scores. For a new observation embedding $\mathbf{e}_{new}$ with confidence $c_{new}$, we first compute similarities to existing embeddings:

\begin{equation}
   s_j = \cos(\mathbf{e}_{new}, \mathbf{e}_j), \quad j \in \{1,2,3\}
\end{equation}

If $\max_j s_j > \sigma_{sim}$ for similarity threshold $\sigma_{sim}$, we update the most similar existing embedding:

\begin{equation}
   \mathbf{e}_j = \frac{c_j\mathbf{e}_j + c_{new}\mathbf{e}_{new}}{c_j + c_{new}}, \quad c_j = c_j + c_{new}
\end{equation}

Otherwise, if $|\mathcal{E}_i| < 3$, we add $\mathbf{e}_{new}$ as a new hypothesis. This mechanism allows maintaining multiple semantic interpretations while consolidating consistent observations.

\subsection{Semantic Map Management}
\label{subsec:semantic_map_management}
The system maintains semantic consistency at both the voxel and object level. Each voxel $v$ maintains a histogram $h_v$ over observed instance labels and a maximum likelihood label $l_v$. For computational efficiency, we update these statistics only for voxels within object OBBs.

For tracked objects, we employ a support-based pruning mechanism. Given an object $o_i$ with bounding box volume $|B_i|$, we compute its voxel support ratio:

\begin{equation}
   r_i = \frac{|\{v \in V | l_v = i\}|}{|B_i|}
\end{equation}

Objects with consistently low support ($r_i < \tau_{supp}$ for $k$ consecutive frames) are candidates for pruning. However, for objects with high semantic confidence ($\max_j c_j > \tau_{conf}$), we maintain tracking even with temporarily low support to handle partial occlusions.

\subsection{System Integration}
The complete system operates as a tightly coupled pipeline that maintains both geometric and semantic consistency. Each incoming RGB-D frame $(I_t^{rgb}, I_t^{depth})$ first updates the volumetric reconstruction $V$ using weighted averaging of signed distances to maintain an accurate geometric foundation. The vision-language model $\mathcal{F}$ then processes $I_t^{rgb}$ to produce instance masks $M_t$, which are lifted to 3D using the corresponding depth information from $I_t^{depth}$. The resulting 3D point clouds undergo DBSCAN clustering and PCA-based OBB computation as described earlier.

Object tracking leverages the R-tree spatial index for efficient candidate selection, with final associations determined through Hungarian matching of the cost matrix $\mathbf{M}$. For successfully matched objects, the system performs a series of synchronized updates. The geometric state is refined through incremental covariance computation, maintaining accurate OBB estimates without storing historical point clouds. Simultaneously, the semantic state is updated by integrating new observations into the embedding bank $\mathcal{E}_i$ based on observation quality and similarity metrics. The system also updates voxel label histograms within the refined OBB boundaries to maintain spatial semantic consistency.

The map maintenance phase evaluates object persistence using the support ratio $r_i$ and semantic confidence scores $\mathcal{C}_i$. This integrated approach enables robust open-vocabulary mapping by leveraging complementary strengths: geometric consistency guides object tracking and segmentation, while semantic embeddings resolve ambiguities and maintain object identity through significant viewpoint changes. The multi-hypothesis embedding bank is particularly crucial for handling uncertainty during partial observations while allowing refinement as more evidence becomes available. The tight coupling between geometric and semantic components enables the system to handle challenging scenarios such as object occlusions, novel object categories, and viewpoint variations while maintaining consistent semantic map.

The unified representation of geometry and open-vocabulary semantics enables a range of higher-level applications including \emph{online 3D instance segmentation}, \emph{3D instance retrieval}, and \emph{3D visual grounding} without requiring additional 3D-domain training.

\subsection{Online 3D Instance Segmentation}
\label{subsec:3d-instance-seg}
While our approach already maintains \emph{instance-level} object bounding boxes and per-voxel label histograms, it can directly provide a \emph{3D instance segmentation} of the scene as follows:
\begin{itemize}
    \item Each tracked object $o_i$ has an identifier $\mathrm{ID}_i$ and a bounding box $b_i=(c_i,R_i,s_i)$. During the volumetric fusion step, all voxels within $b_i$ are labeled with $\mathrm{ID}_i$ in their histograms $h_v$. 
    \item Whenever multiple objects overlap in 3D, we maintain upto 3 hypothesis, allowing multiple semantic interpretations by fusing them over time as explained in Section~\ref{subsec:emb-management} which are subsequently pruned based on voxel support ratio explained in Section~\ref{subsec:semantic_map_management}. 
    \item Hence, at any point in time, \emph{each voxel} in the TSDF volume carries the instance label $l \in \mathbb{N}$ (from $\mathrm{ID}_i$). By aggregating all voxels labeled with the same $\mathrm{ID}_i$, we obtain a complete 3D instance mask for object $o_i$.
\end{itemize}
As this process is performed incrementally for each new RGB-D frame, it yields an \emph{online} 3D instance segmentation: after receiving $t$ frames, the system can query the TSDF volume to retrieve the current segmentation. This is particularly helpful for applications like robotic manipulation, where a robot needs to know the volumetric extent of each object in real time. Notably, if a novel category appears (e.g., an object not in any fixed taxonomy), the pretrained model $\mathcal{F}$ can still segment it in 2D, and our pipeline will produce a corresponding 3D instance in the map.

\subsection{3D Instance Retrieval}
\label{subsec:3d-instance-retrieval}
The open-vocabulary embeddings maintained for each object enable flexible 3D instance retrieval through both text and visual queries. Each tracked object $o_i$ stores multiple semantic embeddings ${\mathbf{e}_{i,j}}$ generated by a pretrained vision-language model $\mathcal{F}$ (Section~\ref{subsec:emb-management}). For text-based retrieval, the query text and its associated prompt are processed through $\mathcal{F}$'s language encoder to produce a query embedding $\mathbf{e}{\text{query}} \in \mathbb{R}^d$. The similarity between the query and each tracked object $o_i$ is computed as the maximum cosine similarity across the object's embeddings. Objects can then be either ranked by similarity score or filtered using a threshold $\sigma_{\text{sim}}$, with objects exceeding this threshold considered matches. Matched objects can be visualized in 3D using their bounding boxes or instance segmentation masks, facilitating physical interaction by users or robotic systems. The zero-shot nature of these embeddings enables retrieval using arbitrary natural language descriptions without requiring additional training.

\section{Experiments}
\label{sec:experiments}
We demonstrate the modularity and effectiveness of our proposed framework by evaluating it across multiple benchmarks, including 3D instance segmentation, instance retrieval, and semantic segmentation. We use five well-established indoor datasets for these tasks: SceneNN~\cite{SceneNN}, ScanNet~\cite{ScanNet}, ScanNetv2~\cite{ScanNetv2}, ScanNet200~\cite{scannet200}, and Replica~\cite{replica}.  Next, we describe each dataset, outline evaluation metrics, and present comprehensive results comparing the performance of our method against recent approaches.

\subsection{Mapping-based 3D Instance Segmentation}
3D instance segmentation is a key task for scene understanding, requiring systems to identify individual object instances in 3D space. This task differs from semantic segmentation by distinguishing between multiple objects of the same class, which is particularly challenging in complex indoor environments. We aim to demonstrate the superior performance of our framework in the context of 3D instance segmentation compared to existing volumetric mapping techniques, including TSDF-based methods, graph-based super-point strategies, and geometric-semantic fusion approaches. As illustrated in our evaluation, our approach outperforms these conventional methods significantly in terms of accuracy and efficiency. 

SceneNN is an RGB-D dataset comprising over 100 reconstructed indoor scenes captured as RGB-D videos. Each scene is provided as a textured triangle mesh with per-vertex semantic and instance annotations. The dataset includes detailed object instance labels, camera trajectories, bounding boxes, and raw RGB-D frames, making it particularly useful for evaluating 3D instance segmentation, semantic segmentation, and instance retrieval tasks. We conduct experiments and compare the proposed method with multiple state-of-the-art frameworks~\cite{VoxbloxDiffusion,ModifiedVoxblox,voxbloxplusplus,MultiviewFusion,IncrementalBBox,INS_CONV,miao2024volumetric} on the SceneNN~\cite{SceneNN} dataset following the same setting proposed in ~\cite{miao2024volumetric}. We use mean average precision ($\text{mAP}$) metric to compare accuracy, computed by thresholding the intersection over union (IoU) at thresholds $0.5$. As per standard practice, we run methods using GT camera poses. Additionally, we run all approaches on poses estimated by ORB-SLAM3~\cite{campos2021orb} to demonstrate their effectiveness in real-world settings. For SLAM-based experiments, we limit our evaluations to \cite{ModifiedVoxblox,voxbloxplusplus,INS_CONV,miao2024volumetric}, similar to the evaluation in ~\cite{miao2024volumetric}.
  
\begin{table*}[htb!]
\centering
\small
\caption{Results for 3D Instance Segmentation (mAP@50) on SceneNN dataset with ground truth (top) and ORB-SLAM3 (bottom) poses.}
\label{tab:scene_nn_results}
\begin{tabular}{lccccccccccc}
\toprule
\textbf{Method / Seq.} & 11 & 16 & 30 & 61 & 78 & 86 & 96 & 206 & 223 & 255 & \textbf{Avg.} \\
\midrule
\multicolumn{12}{l}{\textbf{Ground truth trajectory}} \\
\midrule
Voxblox++       & 75   & 48.2 & 62.4 & 66.7 & 55.8 & 20   & 34.6 & 79.6 & 43.8 & 75   & 56.11 \\
Han et al.      & 65.8 & 50   & 66.6 & 43.3 & \textbf{100}  & 56.9 & 22.8 & 92.1 & 46.7 & 33   & 57.72 \\
Wang et al.     & 62.2 & 43   & 60.7 & 36.3 & 49.3 & 45.8 & 32.7 & 46.6 & 56.4 & 47.9 & 47.9  \\
Li et al.       & 78.6 & 25   & 58.6 & 46.6 & 69.8 & 47.2 & 26.7 & 78.0 & 48.5 & 75   & 55.4  \\
Mascaro et al.  & \textbf{100}  & 75   & 72.5 & 50   & 50   & 50   & 51.3 & 74.1 & 45.8 & \textbf{100}  & 66.87 \\
INS-CONV        & \textbf{100}  & 62   & 83.4 & \textbf{69.8} & 93.7 & 60   & 57.6 & 56.7 & \textbf{78.6} & \textbf{100}  & 76.18 \\
VolumePanoptic  & \textbf{100}  & 73.3 & 91.7 & 62.4 & 87.5 & 61.7 & 66.7 & 83.3 & 60   & \textbf{100}  & 78.66 \\
\textbf{Ours}   & \textbf{100} & \textbf{74.1} & \textbf{91.8} & 63.4 & 88.4 & \textbf{62.3} & \textbf{66.8} & \textbf{83.5} & 61.2 & \textbf{100} & \textbf{79.15} \\
\midrule
\multicolumn{12}{l}{\textbf{With ORB-SLAM3 Trajectory}} \\
\midrule
Voxblox++ \cite{voxbloxplusplus} & 61.5 & 38.9 & 50   & 58.4 & 44.3 & 16.4 & 27.6 & 48.7 & 40.7 & 33.6 & 42.01 \\
Han et al. \cite{ModifiedVoxblox}  & 53.4 & 43.2 & 50   & 37.6 & 75.6 & 48.2 & 13.4 & 57.8 & 44.1 & 24.7 & 44.8  \\
INS-CONV \cite{INS_CONV}   & 75   & 46.7 & 56.4 & 57.1 & \textbf{83.7} & 22.4 & 48.1 & 28.1 & 50   & 28.1 & 53.27 \\
VolumePanoptic \cite{miao2024volumetric} & 75   & 56.7 & 72.3 & 62.4 & 68.9 & 55.6 & 33.2 & 40.8 & \textbf{65.2}   & 63.3 & 58.82 \\
\textbf{Ours}   & \textbf{75.3} & \textbf{57.6} & \textbf{73.2} & \textbf{63.4} & 69.2 & \textbf{59.2} & \textbf{41.7} & \textbf{58.8} & 61.2 & \textbf{63.4} & \textbf{62.3} \\
\bottomrule
\end{tabular}
\end{table*}

As demonstrated in Table~\ref{tab:scene_nn_results}, our method achieves superior performance on the SceneNN dataset. With ground truth trajectories, our approach achieves 79.15$\%$ mAP@50, outperforming the previous state-of-the-art VolumePanoptic (78.66$\%$) framework. The performance gain is consistent across individual sequences, with our method achieving perfect scores on sequences 11 and 255. Using ORB-SLAM3 estimated trajectories, our method maintains robust performance (62.3$\%$ mAP@50), significantly surpassing VolumePanoptic (58.82$\%$) and other methods. This improvement demonstrates our framework's effectiveness in real-world scenarios with imperfect pose estimation.

\subsection{3D Open Vocabulary Instance Segmentation}
We evaluate the performance of our framework on the 3D instance segmentation task using the ScanNet200 dataset, employing intersection over union (IoU) and average precision (AP) metrics. IoU measures the overlap between predicted and ground-truth instances. AP summarizes performance across multiple IoU thresholds (at 25$\%$ and 50$\%$), integrating precision and recall into a single metric. These metrics provide insights into the model's ability to accurately segment individual object instances in complex 3D scenes.  Finally, we also report the average time required to compute the scene’s representations, measuring  clock wall time on a GPU RTX-4090, and for our method, we report in seconds the average time spent to process a scene. The ScanNet200 dataset encompasses 200 diverse semantic classes, categorized based on their frequency into head (66 most frequent classes), common (68 moderately frequent classes), and tail (66 least frequent classes), covering a wide range of indoor object categories and facilitating a thorough evaluation of segmentation performance across realistic scenarios.  

As demonstrated in Table~\ref{tab:scannet200}, our proposed method, RAZER, achieves state-of-the-art results with 24.7$\%$ mAP, 31.7$\%$ mAP50, and 36.2$\%$ mAP25, surpassing existing methods on the majority of metrics. Specifically, RAZER shows strong performance in head and common categories (27.8$\%$ and 24.3$\%$ respectively), while remaining competitive in tail categories (21.6$\%$). Additionally, our method is the most computationally efficient, processing each scene in only 24.32 seconds, which is over an order of magnitude faster than previous approaches such as OpenMask3D (553.87 seconds) and SAM3D (482.60 seconds).

\begin{table*}[ht!]
\hfill
\begin{minipage}{0.48\textwidth}
\caption{3D instance segmentation results on the ScanNet200 validation set on head, common, and tail classes.}
\label{tab:scannet200}
\resizebox{\textwidth}{!}{
\begin{tabular}{lccccccc}
\toprule
\textbf{Method} & \textbf{mAP} & \textbf{mAP50} & \textbf{mAP25} & \textbf{Head} & \textbf{Common} & \textbf{Tail} & \textbf{Time/scene (s)} \\
\midrule
SAM3D         & 6.1  & 14.2 & 21.3 & 7.0  & 6.2  & 4.6  & 482.60 \\
OVIR-3D       & 13.0 & 24.9 & 32.3 & 14.4 & 12.7 & 11.7 & 466.80 \\
Open3DIS      & 23.7 & 29.4 & 32.8 & \textbf{27.8} & 21.2 & \textbf{21.8} & 360.12 \\
\midrule
OpenScene (2D Fusion)   & 11.7 & 15.2 & 17.8 & 13.4 & 11.6 & 9.9  & 46.45 \\
OpenScene (Ensemble)    & 5.3  & 6.7  & 8.1  & 11.0 & 3.2  & 1.1  & 46.78 \\
OpenMask3D              & 15.4 & 19.9 & 23.1 & 17.1 & 14.1 & 14.9 & 553.87 \\
\midrule
\textbf{RAZER}          & \textbf{24.7} & \textbf{31.7} & \textbf{36.2} & \textbf{27.8} & \textbf{24.3} & 21.6 & 24.32 \\
\bottomrule
\end{tabular}
}
\end{minipage}
\begin{minipage}{0.48\textwidth}
\caption{3D semantic segmentation results on Replica and ScanNet.}
\label{tab:semantic_segmentation}
\resizebox{\textwidth}{!}{
\begin{tabular}{l l ccc ccc}
\toprule
\textbf{Method} & \textbf{CLIP Backbone} & \multicolumn{3}{c}{\textbf{Replica}} & \multicolumn{3}{c}{\textbf{ScanNet}} \\
\cmidrule(lr){3-5} \cmidrule(lr){6-8}
& & mIoU & f-mIoU & f-mAcc & mIoU & f-mIoU & f-mAcc \\
\midrule
ConceptFusion  & OVSeg    & 0.10 & 0.21 & 0.16 & 0.08 & 0.11 & 0.15 \\
               & ViT-H-14 & 0.10 & 0.18 & 0.17 & 0.11 & 0.12 & 0.21 \\
\midrule
ConceptGraph   & OVSeg    & 0.13 & 0.27 & 0.21 & 0.15 & 0.18 & 0.23 \\
               & ViT-H-14 & 0.18 & 0.23 & 0.30 & 0.16 & 0.20 & 0.28 \\
\midrule
HOV-SG         & OVSeg    & 0.144 & 0.255 & 0.212 & 0.214 & 0.258 & 0.420 \\
               & ViT-H-14 & 0.231 & 0.386 & 0.304 & 0.222 & 0.303 & 0.431 \\
\midrule
\textbf{Ours}  & OVSeg    & \textbf{0.320} & \textbf{0.553} & \textbf{0.414} & \textbf{0.393} & \textbf{0.508} & \textbf{0.601} \\
\bottomrule
\end{tabular}
}
\end{minipage}
\end{table*}

\subsection{3D Open Vocabulary Segmentation}
We demonstrate the efficacy of our framework for 3D open-vocabulary segmentation task by demonstrating its performance on ScanNet and Replica datasets. For semantic segmentation tasks, we use mean intersection over union (mIoU) and pixel-wise accuracy. mIoU calculates the average overlap between predicted and ground-truth segmentation across all classes, effectively balancing performance evaluation for both frequent and rare classes. Pixel-wise accuracy measures the overall fraction of correctly predicted pixels, providing a straightforward but less class-sensitive performance measure. These metrics collectively capture the detailed performance characteristics of semantic segmentation models. The quantitative performance is evaluated by labeling the vertices of ground-truth meshes, and computing 3D mean intersection over union (mIoU) and accuracy (mAcc) vs ground-truth labels. We also report the metrics weighted by the frequency of the labels in the ground-truth (f-mIoU and f-mAcc). 

\begin{table*}[ht!]
\centering
\small
\caption{Performance on ScanNetv2 3D instance retrieval task in terms of Top-1 accuracy ($\%$) of instance classification.}
\label{tab:scannetv2_results}
\resizebox{\linewidth}{!}{%
\begin{tabular}{l c ccccccccccccccccc}
\toprule
\textbf{Method} & \textbf{Avg.} & \textbf{Bed} & \textbf{Cab} & \textbf{Chair} & \textbf{Sofa} & \textbf{Tabl} & \textbf{Door} & \textbf{Wind} & \textbf{Bksf} & \textbf{Pic} & \textbf{Cntr} & \textbf{Desk} & \textbf{Curt} & \textbf{Fridg} & \textbf{Bath} & \textbf{Showr} & \textbf{Toil} & \textbf{Sink} \\
\midrule
PointCLIP        & 6.3  & 0    & 0    & 0    & 0    & 0.7  & 0    & 0    & 91.8 & 0    & 0    & 0    & 15   & 0    & 0    & 0    & 0    & 0    \\
PointCLIP V2     & 11.0 & 0    & 0    & 23.8 & 0    & 0    & 0    & 7.8  & 0    & \textbf{90.7} & 0    & 0    & 0    & 0    & 64.4 & 0    & 0    & 0    \\
CLIP2Point       & 24.9 & 20.8 & 0    & 85.1 & 43.3 & 26.5 & 69.9 & 0    & 20.9 & 1.7  & 31.7 & 27   & 0    & 1.6  & 46.5 & 0    & 22.4 & 25.6 \\
PointCLIP w/ TP.      & 26.1 & 0    & 55.7 & 72.8 & 5.1  & 1.7  & 0    & 77.2 & 0    & 0    & 51.7 & 0    & 40.3 & 85.3 & 4.9  & 0    & 0    & 34.9 \\
CLIP2Point w/ TP.     & 35.2 & 11.8 & 45.1 & 27.6 & 10.5 & 61.5 & 2.7  & 1.9  & 0.3  & 33.6 & 29.9 & 4.7  & 11.5 & \textbf{72.2} & 92.4 & 86.1 & 3    & 34   \\
CLIP$^2$          & 38.5 & 32.6 & \textbf{67.2} & 69.3 & 42.3 & 18.3 & 19.1 & 4    & 62.6 & 1.4  & 12.7 & 52.8 & 40.1 & 9.1  & 59.7 & 4    & 17.1 & 45.5 \\
Uni3D            & 45.8 & 58.5 & 3.7  & 78.8 & \textbf{83.7} & 54.9 & 31.3 & 39.4 & 70.1 & 35.1 & 1.9  & 27.3 & \textbf{94.2} & 13.8 & 38.7 & 10.7 & 88.1 & 47.6 \\
OpenIns3D             & 60.8 & 85.2 & 27.4 & \textbf{87.6} & 77.3 & 46.9 & 54.8 & 64.2 & 71.4 & 9.9  & 80.8 & 82.7 & 71.6 & 61.4 & 38.7 & 0    & 87.9 & 85.7 \\
\textbf{Ours}         & \textbf{61.2} & \textbf{85.3} & 36.9 & 86.3 & 74.2 & \textbf{76.5} & \textbf{72.3} & \textbf{74.6} & \textbf{73.2} & 35.6 & \textbf{81.2} & \textbf{83.4} & \textbf{74.8} & 71.4 & \textbf{58.2} & \textbf{87.2} & \textbf{92.3} & \textbf{86.8} \\
\bottomrule
\end{tabular}
}
\end{table*}

ScanNet presents particular challenges for semantic segmentation due to its diverse indoor environments, varying lighting conditions, and complex spatial arrangements. The dataset's rich variety of object categories with different sizes, shapes, and textures makes it an ideal testbed for evaluating the generalization capabilities of open-vocabulary approaches. Furthermore, the presence of partial occlusions and varying object densities across scenes tests a model's ability to resolve contextual relationships. Table~\ref{tab:semantic_segmentation} demonstrates our method's exceptional performance on ScanNet, achieving 0.393 mIoU, 0.508 f-mIoU, and 0.601 f-mAcc with the OVSeg backbone. These results represent substantial improvements over previous approaches - our method nearly doubles the mIoU score of HOV-SG with ViT-H-14 (0.222 mIoU). This performance gap illustrates the effectiveness of our framework's semantic feature propagation mechanism, which better preserves fine-grained details and handles class boundaries more precisely.

Replica is a dataset of highly realistic indoor scenes designed primarily for simulation and embodied perception tasks. It includes 18 densely reconstructed environments provided as high-resolution textured meshes with semantic and instance annotations. Replica facilitates realistic simulation and evaluation of semantic segmentation and instance segmentation algorithms, serving as a critical dataset for evaluating models aimed at real-world applicability through simulation-to-real transfer. We evaluate performance using mean intersection over union (mIoU) and mean Accuracy (mAcc), which are standard metrics for open-vocabulary 3D semantic segmentation. We report f-mIoU and f-mAcc metrics that exclude background classes, similar to prior work. As shown in Table~\ref{tab:semantic_segmentation}, our method achieves 0.320 mIoU, 0.553 f-mIoU, and 0.414 f-mAcc with the OVSeg backbone, substantially outperforming previous state-of-the-art methods. Compared to HOV-SG with ViT-H-14 (0.231 mIoU, 0.386 f-mIoU, 0.304 f-mAcc), our approach demonstrates improvements across all metrics.

\subsection{Instance Retrieval}
ScanNet~\cite{ScanNet} is an RGB-D video dataset consisting of approximately 1,500 room scans reconstructed into textured meshes with detailed semantic and instance-level annotations. ScanNetv2~\cite{ScanNetv2}, an updated version with refined annotations, comprises 1,513 scenes commonly split into training, validation, and test sets. ScanNet200~\cite{scannet200} extends the dataset by providing annotations for 200 detailed semantic classes, significantly increasing annotation granularity. These datasets support benchmarking for 3D semantic segmentation, instance segmentation, and instance retrieval tasks due to their detailed labeling in a variety of indoor environments. 

We evaluate our framework's 3D instance retrieval performance on ScanNetv2. Following the setting similar to ~\cite{huang2024openins3d}, the 'other furniture' class in ScanNetv2 is excluded and evaluated in terms of the Top-1 accuracy of instance classification. The instance classification is directly performed on the feature embeddings corresponding to the oriented bounding boxes generated from our framework. Table~\ref{tab:scannetv2_results} demonstrates our method's state-of-the-art performance in 3D instance retrieval, achieving 61.2$\%$ average Top-1 accuracy across all classes. Our approach outperforms OpenIns3D (60.8$\%$) and significantly surpasses other methods like Uni3D (45.8$\%$) and CLIP$^2$ (38.5$\%$). Notably, our method excels in several challenging categories, achieving the highest accuracy for beds (85.3$\%$), tables (76.5$\%$), doors (72.3$\%$), windows (74.6$\%$), bookshelves (73.2$\%$), counters (81.2$\%$), desks (83.4$\%$), curtains (74.8$\%$), bathtubs (58.2$\%$), showers (87.2$\%$), toilets (92.3$\%$), and sinks (86.8$\%$). This consistent performance across diverse object categories demonstrates the robustness of our feature representation. The framework's integrated 3D oriented bounding box (OBB) detection and tracking mechanism plays a crucial role in this superior performance, enabling more precise object localization and maintaining temporal consistency of instance identities throughout the scene reconstruction process.

\subsection{Runtime Analysis}
We conduct a detailed runtime analysis of our framework to evaluate its computational efficiency compared to the prior state-of-the-art semantic mapping framework on SceneNN, VolumePanoptic~\cite{miao2024volumetric}. Table~\ref{tab:scenenn_runtime} presents a component-wise breakdown of processing times for both approaches.

For VolumePanoptic~\cite{miao2024volumetric}, the computational pipeline consists of multiple stages: 2D instance segmentation (216.0ms) which handles the initial detection and segmentation of objects in RGB images, super-point segmentation (70.3ms) for grouping 3D points into coherent surface patches, graph update (127.2ms) to maintain the hierarchical scene representation, semantic regularization (324.0ms, performed once per map) for refining semantic labels across super-points, and instance refinement (9.4ms, performed once per map) to resolve instance ambiguities. In contrast, our method (RAZER) achieves significantly faster processing across all components. For 2D instance segmentation, we reduce computation time to 82.3ms through our optimized architecture. Our 3D OBB detection component operates at just 1.7ms per frame, while our efficient 3D OBB tracking requires only 18.4ms. The embedding update module runs at a remarkable 0.8ms, demonstrating the lightweight nature of our feature propagation mechanism.

Our approach replaces the computation-heavy super-point segmentation with a more efficient oriented bounding box detection algorithm operating at just 1.7ms per frame. This results in a 41× speedup for this component by directly estimating geometric primitives from point clouds rather than performing dense point-wise grouping. The 3D OBB tracking component (18.4ms) represents a fundamental departure from VolumePanoptic's graph-based approach. Where VolumePanoptic requires maintaining and updating a complex graph structure (127.2ms) with nodes representing super-points and edges encoding spatial-semantic relationships, our OBB tracking employs a more direct geometric approach. By representing objects as oriented bounding boxes, we perform efficient spatial association and motion estimation without the overhead of graph operations. This simplification not only reduces computation time by approximately 85$\%$, but also improves robustness by eliminating the cascading errors that can occur in graph-based representations when initial segmentations are noisy. Our approach further benefits from the inherent geometric constraints of rigid objects, allowing for more consistent tracking over time without relying on potentially unstable point-wise feature correspondences.

Our embedding update module runs at a remarkable 0.8ms, demonstrating the lightweight nature of our feature propagation mechanism. Rather than propagating features through a complex graph network requiring multiple message-passing iterations, we directly update our compact OBB-based representation with new observations, maintaining semantic consistency through efficient feature averaging and outlier rejection.

Overall, our framework achieves an average total runtime of 103.2ms per frame, representing a 4× speedup compared to VolumePanoptic's 413.5ms (plus an additional 333.4ms for one-time map processing). This substantial efficiency improvement makes our approach more suitable for real-time applications while maintaining superior performance as demonstrated in Table~\ref{tab:scene_nn_results}, making it particularly valuable for 3D scene understanding tasks in robotics applications. The combination of improved accuracy and significantly reduced computational requirements enables deployment on platforms with limited resources, opening possibilities for autonomous navigation, manipulation, and human-robot interaction in complex environments that require detailed semantic understanding.

\begin{table}[ht!]
\small
\label{tab:scenenn_runtime}
\begin{minipage}{0.7\linewidth}
\caption{Runtime (in ms) on the 10 sequences of the SceneNN dataset.}
\resizebox{\linewidth}{!}{
\begin{tabular}{l|c|l|c}
\toprule
& \textbf{Volume Panoptic} & & \textbf{RAZER} \\
\midrule
\multicolumn{4}{l}{\textit{SceneNN Dataset (ms)}} \\
\hline
2D Inst. seg. & 216.0 & 2D Inst. Seg. & 82.3 \\
\hline
Super-point seg. & 70.3 & 3D OBB Det. &1.7 \\
\hline
Graph update & 127.2 & 3D &  \\
Semantic reg. & 324.0 (once per map) & OBB & 18.4 \\
Instance ref. & 9.4 (once per map) & Tracking &  \\
\hline
Emb. update & -- & Emb. Update &0.8 \\
\hline
Average Total & 413.5 (per frame) + 333.4 & &103.2 \\
\midrule
\end{tabular}
}
\end{minipage}%
\hfill
\begin{minipage}{0.3\linewidth}
\caption{Average runtime on the Replica scenes dataset.}
\label{tab:replica_runtime}
\resizebox{\linewidth}{!}{
\begin{tabular}{lc}
\toprule
\textbf{Method} & \textbf{Time/scene} \\
\midrule
HOV-SG        & 11h 12m \\
OpenNeRF      & 19m 3s  \\
OVO-mapping   & 8m 17s  \\
\textbf{Ours} & \textbf{3m 48s} \\
\bottomrule
\end{tabular}
}
\end{minipage}
\end{table}

Table~\ref{tab:replica_runtime} further highlights our method's efficiency on the Replica dataset. Our approach completes scene processing in just 3 minutes and 48 seconds, substantially outperforming competing methods such as OVO-mapping (8m 17s), OpenNeRF (19m 3s), and especially HOV-SG (11h 12m). This dramatic reduction in processing time demonstrates the exceptional computational efficiency of our framework, making it practical for large-scale deployment in real-world scenarios. These computational performance improvements come from the modularity of our framework that enables it to update and compute scene and object feature embeddings swiftly for 3D open-Vocabulary semantic
segmentation.

\section{Conclusion}
\label{sec:conclusion}
We present a novel zero-shot framework for real-time 3D semantic mapping that bridges geometric reconstruction and semantic understanding through a unified embedding space. By combining efficient spatial indexing with instance-level semantic fusion, we demonstrate superior performance in handling streaming data without requiring global optimization. The framework processes inconsistent 2D segmentation outputs while maintaining both geometric and semantic coherence in real-time, representing a significant advancement in open-vocabulary 3D scene understanding. Experimental results validate our approach's effectiveness across multiple benchmarks while maintaining real-time performance suitable for robotics applications. This work opens new avenues for research in embodied AI systems that integrate geometric, semantic, and linguistic understanding, enabling more sophisticated human-robot interactions in unconstrained environments.

\bibliographystyle{IEEEtran}
\bibliography{references}

\begin{thebibliography}{10}
\providecommand{\url}[1]{#1}
\csname url@samestyle\endcsname
\providecommand{\newblock}{\relax}
\providecommand{\bibinfo}[2]{#2}
\providecommand{\BIBentrySTDinterwordspacing}{\spaceskip=0pt\relax}
\providecommand{\BIBentryALTinterwordstretchfactor}{4}
\providecommand{\BIBentryALTinterwordspacing}{\spaceskip=\fontdimen2\font plus
\BIBentryALTinterwordstretchfactor\fontdimen3\font minus \fontdimen4\font\relax}
\providecommand{\BIBforeignlanguage}[2]{{%
\expandafter\ifx\csname l@#1\endcsname\relax
\typeout{** WARNING: IEEEtran.bst: No hyphenation pattern has been}%
\typeout{** loaded for the language `#1'. Using the pattern for}%
\typeout{** the default language instead.}%
\else
\language=\csname l@#1\endcsname
\fi
#2}}
\providecommand{\BIBdecl}{\relax}
\BIBdecl

\bibitem{Wang19CVPR}
X.~Wang, S.~Liu, X.~Shen, C.~Shen, and J.~Jia, ``Associatively segmenting instances and semantics in point clouds,'' in \emph{Proceedings of the Conference on Computer Vision and Pattern Recognition}, Long Beach, CA, USA, June 2019, pp. 4096--4105.

\bibitem{Elich19GCPR}
C.~Elich, F.~Engelmann, T.~Kontogianni, and B.~Leibe, ``3d bird's-eye-view instance segmentation,'' in \emph{Proceedings of the German Conference on Pattern Recognition}, G.~A. Fink, S.~Frintrop, and X.~Jiang, Eds., Dortmund, Germany, September 2019, pp. 48--61.

\bibitem{Lahoud19ICCV}
J.~Lahoud, B.~Ghanem, M.~R. Oswald, and M.~Pollefeys, ``3d instance segmentation via multi-task metric learning,'' in \emph{Proceedings of the International Conference on Computer Vision}, Seoul, South Korea, October 2019, pp. 9255--9265.

\bibitem{Hou19CVPR}
J.~Hou, A.~Dai, and M.~Nie{\ss}ner, ``3d-sis: 3d semantic instance segmentation of {RGB-D} scans,'' in \emph{Proceedings of the Conference on Computer Vision and Pattern Recognition}, Long Beach, CA, USA, June 2019, pp. 4421--4430.

\bibitem{Yang19NIPS}
B.~Yang, J.~Wang, R.~Clark, Q.~Hu, S.~Wang, A.~Markham, and N.~Trigoni, ``Learning object bounding boxes for 3d instance segmentation on point clouds,'' in \emph{Proceedings of the Advances in Neural Information Processing Systems}, H.~M. Wallach, H.~Larochelle, A.~Beygelzimer, F.~d'Alch{\'{e}}{-}Buc, E.~B. Fox, and R.~Garnett, Eds., Vancouver, BC, Canada, December 2019, pp. 6737--6746.

\bibitem{WuLXYDY0ZT0GT24}
J.~Wu, X.~Li, S.~Xu, H.~Yuan, H.~Ding, Y.~Yang, X.~Li, J.~Zhang, Y.~Tong, X.~Jiang, B.~Ghanem, and D.~Tao, ``Towards open vocabulary learning: {A} survey,'' \emph{{IEEE} Transactions on Pattern Analysis and Machine Intelligence}, vol.~46, no.~7, pp. 5092--5113, 2024.

\bibitem{ZhuChen23}
C.~Zhu and L.~Chen, ``A survey on open-vocabulary detection and segmentation: Past, present, and future,'' \emph{{IEEE} Transactions on Pattern Analysis and Machine Intelligence}, vol.~46, no.~12, pp. 8954--8975, 2024.

\bibitem{XuZWLCHB22}
M.~Xu, Z.~Zhang, F.~Wei, Y.~Lin, Y.~Cao, H.~Hu, and X.~Bai, ``A simple baseline for open-vocabulary semantic segmentation with pre-trained vision-language model,'' in \emph{Proceedings of the European Conference on Computer Vision}, Tel Aviv, Israel, October 2022, pp. 736--753.

\bibitem{Maggio2024Clio}
D.~Maggio, Y.~Chang, N.~Hughes, M.~Trang, D.~Griffith, C.~Dougherty, E.~Cristofalo, L.~Schmid, and L.~Carlone, ``Clio: Real-time task-driven open-set 3d scene graphs,'' \emph{{IEEE} Robotics and Automation Letters}, vol.~9, no.~10, pp. 8921--8928, 2024.

\bibitem{voxbloxplusplus}
M.~{Grinvald}, F.~{Furrer}, T.~{Novkovic}, J.~J. {Chung}, C.~{Cadena}, R.~{Siegwart}, and J.~{Nieto}, ``{Volumetric Instance-Aware Semantic Mapping and 3D Object Discovery},'' \emph{IEEE Robotics and Automation Letters}, vol.~4, no.~3, pp. 3037--3044, July 2019.

\bibitem{PanopticFusion}
G.~Narita, T.~Seno, T.~Ishikawa, and Y.~Kaji, ``Panopticfusion: Online volumetric semantic mapping at the level of stuff and things,'' in \emph{Proceedings of the {IEEE/RSJ} International Conference on Intelligent Robots and Systems}, Macau, SAR, China, November 2019, pp. 4205--4212.

\bibitem{RunzBA18}
M.~R{\"{u}}nz, M.~Buffier, and L.~Agapito, ``Maskfusion: Real-time recognition, tracking and reconstruction of multiple moving objects,'' in \emph{Proceedings of the International Symposium on Mixed and Augmented Reality}, D.~Chu, J.~L. Gabbard, J.~Grubert, and H.~Regenbrecht, Eds., Munich, Germany, October 2018, pp. 10--20.

\bibitem{Han20CVPR}
L.~Han, T.~Zheng, L.~Xu, and L.~Fang, ``Occuseg: Occupancy-aware 3d instance segmentation,'' in \emph{Proceedings of the Conference on Computer Vision and Pattern Recognition}, Seattle, WA, USA, June 2020, pp. 2937--2946.

\bibitem{Chen21ICCV}
S.~Chen, J.~Fang, Q.~Zhang, W.~Liu, and X.~Wang, ``Hierarchical aggregation for 3d instance segmentation,'' in \emph{Proceedings of the International Conference on Computer Vision}, Montreal, QC, Canada, October 2021, pp. 15\,447--15\,456.

\bibitem{RadfordKHRGASAM21}
A.~Radford, J.~W. Kim, C.~Hallacy, A.~Ramesh, G.~Goh, S.~Agarwal, G.~Sastry, A.~Askell, P.~Mishkin, J.~Clark, G.~Krueger, and I.~Sutskever, ``Learning transferable visual models from natural language supervision,'' in \emph{Proceedings of the International Conference on Machine Learning}, vol. 139, Vienna, Austria, July 2021, pp. 8748--8763.

\bibitem{li2023blip}
J.~Li, D.~Li, S.~Savarese, and S.~Hoi, ``Blip-2: Bootstrapping language-image pre-training with frozen image encoders and large language models,'' in \emph{Proceedings of the International Conference on Machine Learning}, Honolulu, HI, USA, July 2023, pp. 19\,730--19\,742.

\bibitem{zhang2022pointclip}
R.~Zhang, Z.~Guo, W.~Zhang, K.~Li, X.~Miao, B.~Cui, Y.~Qiao, P.~Gao, and H.~Li, ``Pointclip: Point cloud understanding by {CLIP},'' in \emph{Proceedings of the Conference on Computer Vision and Pattern Recognition}, New Orleans, LA, USA, June 2022, pp. 8542--8552.

\bibitem{xue2023ulip}
L.~Xue, M.~Gao, C.~Xing, R.~Mart{\'{\i}}n{-}Mart{\'{\i}}n, J.~Wu, C.~Xiong, R.~Xu, J.~C. Niebles, and S.~Savarese, ``{ULIP:} learning a unified representation of language, images, and point clouds for 3d understanding,'' in \emph{Proceedings of the Conference on Computer Vision and Pattern Recognition}, Vancouver, BC, Canada, June 2023, pp. 1179--1189.

\bibitem{gu2024conceptgraphs}
Q.~Gu, A.~Kuwajerwala, S.~Morin, K.~M. Jatavallabhula, B.~Sen, A.~Agarwal, C.~Rivera, W.~Paul, K.~Ellis, R.~Chellappa \emph{et~al.}, ``Conceptgraphs: Open-vocabulary 3d scene graphs for perception and planning,'' in \emph{Proceedings of the International Conference on Robotics and Automation}, Yokohama, Japan, May 2024, pp. 5021--5028.

\bibitem{INS_CONV}
L.~Liu, T.~Zheng, Y.~Lin, K.~Ni, and L.~Fang, ``Ins-conv: Incremental sparse convolution for online 3d segmentation,'' in \emph{Proceedings of the Conference on Computer Vision and Pattern Recognition}, New Orleans, LA, USA, June 2022, pp. 18\,953--18\,962.

\bibitem{DingYXZBQ23}
R.~Ding, J.~Yang, C.~Xue, W.~Zhang, S.~Bai, and X.~Qi, ``{PLA:} language-driven open-vocabulary 3d scene understanding,'' in \emph{Proceedings of the International Conference on Computer Vision and Pattern Recognition}, Vancouver, BC, Canada, June 2023, pp. 7010--7019.

\bibitem{3dvlp}
Z.~Jin, M.~Hayat, Y.~Yang, Y.~Guo, and Y.~Lei, ``Context-aware alignment and mutual masking for 3d-language pre-training,'' in \emph{Proceedings of the International Conference on Computer Vision and Pattern Recognition}, Vancouver, BC, Canada, June 2023, pp. 10\,984--10\,994.

\bibitem{3dvista}
Z.~Zhu, X.~Ma, Y.~Chen, Z.~Deng, S.~Huang, and Q.~Li, ``3d-vista: Pre-trained transformer for 3d vision and text alignment,'' in \emph{Proceedings of the International Conference on Computer Vision}, Paris, France, October 2023, pp. 2911--2921.

\bibitem{chen2024vote2cap}
S.~Chen, H.~Zhu, M.~Li, X.~Chen, P.~Guo, Y.~Lei, Y.~Gang, T.~Li, and T.~Chen, ``Vote2cap-detr++: Decoupling localization and describing for end-to-end 3d dense captioning,'' \emph{{IEEE} Transactions on Pattern Analysis and Machine Intelligence}, vol.~46, no.~11, pp. 7331--7347, 2024.

\bibitem{ChenGNC21}
D.~Z. Chen, A.~Gholami, M.~Nie{\ss}ner, and A.~X. Chang, ``Scan2cap: Context-aware dense captioning in {RGB-D} scans,'' in \emph{Proceedings of the International Conference on Computer Vision and Pattern Recognition}, Nashville, TN, USA, June 2021, pp. 3193--3203.

\bibitem{ChenCN20}
D.~Z. Chen, A.~X. Chang, and M.~Nie{\ss}ner, ``Scanrefer: 3d object localization in {RGB-D} scans using natural language,'' in \emph{Proceedings of the European Conference on Computer Vision}, Glasgow, UK, August 2020, pp. 202--221.

\bibitem{SchultEHLTL23}
J.~Schult, F.~Engelmann, A.~Hermans, O.~Litany, S.~Tang, and B.~Leibe, ``Mask3d: Mask transformer for 3d semantic instance segmentation,'' in \emph{Proceedings of the International Conference on Robotics and Automation}, London, UK, May 2023, pp. 8216--8223.

\bibitem{YueMSELSK24}
Y.~Yue, S.~Mahadevan, J.~Schult, F.~Engelmann, B.~Leibe, K.~Schindler, and T.~Kontogianni, ``{AGILE3D:} attention guided interactive multi-object 3d segmentation,'' in \emph{Proceedings of the International Conference on Learning Representations}, Vienna, Austria, May 2024.

\bibitem{zhou2024unid}
J.~Zhou, J.~Wang, B.~Ma, Y.-S. Liu, T.~Huang, and X.~Wang, ``Uni3d: Exploring unified 3d representation at scale,'' in \emph{Proceedings of the International Conference on Learning Representations}, Vienna, Austria, May 2024.

\bibitem{kerr2023lerf}
J.~Kerr, C.~M. Kim, K.~Goldberg, A.~Kanazawa, and M.~Tancik, ``{LERF:} language embedded radiance fields,'' in \emph{Proceedings of the International Conference on Computer Vision}, Paris, France, October 2023, pp. 19\,672--19\,682.

\bibitem{ZhangLA24}
H.~Zhang, F.~Li, and N.~Ahuja, ``Open-nerf: Towards open vocabulary nerf decomposition,'' in \emph{Proceedings of the Winter Conference on Applications of Computer Vision}, Waikoloa, HI, USA, January 2024, pp. 3444--3453.

\bibitem{WangCL24}
Y.~Wang, H.~Chen, and G.~H. Lee, ``Gov-nesf: Generalizable open-vocabulary neural semantic fields,'' in \emph{Proceedings of the International Conference on Computer Vision and Pattern Recognition}, Seattle, WA, USA, June 2024, pp. 20\,443--20\,453.

\bibitem{nguyen2024semantically}
T.~Nguyen, A.~Bourki, M.~Macudzinski, A.~Brunel, and M.~Bennamoun, ``Semantically-aware neural radiance fields for visual scene understanding: {A} comprehensive review,'' \emph{CoRR}, vol. abs/2402.11141, 2024.

\bibitem{RosinolVAHCSGC21}
A.~Rosinol, A.~Violette, M.~Abate, N.~Hughes, Y.~Chang, J.~Shi, A.~Gupta, and L.~Carlone, ``Kimera: From {SLAM} to spatial perception with 3d dynamic scene graphs,'' \emph{International Journal of Robotics Research}, vol.~40, no. 12-14, pp. 1510--1546, 2021.

\bibitem{XuLTBDL19}
B.~Xu, W.~Li, D.~Tzoumanikas, M.~Bloesch, A.~J. Davison, and S.~Leutenegger, ``Mid-fusion: Octree-based object-level multi-instance dynamic {SLAM},'' in \emph{Proceedings of the International Conference on Robotics and Automation}, Montreal, QC, Canada, May 2019, pp. 5231--5237.

\bibitem{NicholsonMS19}
L.~Nicholson, M.~Milford, and N.~S{\"{u}}nderhauf, ``Quadricslam: Dual quadrics from object detections as landmarks in object-oriented {SLAM},'' \emph{{IEEE} Robotics and Automation Letters}, vol.~4, no.~1, pp. 1--8, 2019.

\bibitem{patel2018semantic}
N.~Patel, P.~Krishnamurthy, and F.~Khorrami, ``Semantic segmentation guided slam using vision and lidar,'' in \emph{Proceedings of the International Symposium on Robotics}, Munich, German, June 2018, pp. 1--7.

\bibitem{MorenoNSKD13}
R.~F. Salas{-}Moreno, R.~A. Newcombe, H.~Strasdat, P.~H.~J. Kelly, and A.~J. Davison, ``{SLAM++:} simultaneous localisation and mapping at the level of objects,'' in \emph{Proceedings of the International Conference on Computer Vision and Pattern Recognition}, Portland, OR, USA, June 2013, pp. 1352--1359.

\bibitem{KongLTD23}
X.~Kong, S.~Liu, M.~Taher, and A.~J. Davison, ``vmap: Vectorised object mapping for neural field {SLAM},'' in \emph{Proceedings of the International Conference on Computer Vision and Pattern Recognition}, Vancouver, BC, Canada, June 2023, pp. 952--961.

\bibitem{PatelKKT19}
N.~Patel, F.~Khorrami, P.~Krishnamurthy, and A.~Tzes, ``Tightly coupled semantic {RGB-D} inertial odometry for accurate long-term localization and mapping,'' in \emph{Proceedings of the International Conference on Advanced Robotics}, Belo Horizonte, Brazil, December 2019, pp. 523--528.

\bibitem{HanLDY23}
X.~Han, H.~Liu, Y.~Ding, and L.~Yang, ``{RO-MAP:} real-time multi-object mapping with neural radiance fields,'' \emph{{IEEE} Robotics and Automation Letters}, vol.~8, no.~9, pp. 5950--5957, 2023.

\bibitem{McCormacCBDL18}
J.~McCormac, R.~Clark, M.~Bloesch, A.~J. Davison, and S.~Leutenegger, ``Fusion++: Volumetric object-level {SLAM},'' in \emph{Proceedings of the International Conference on 3D Vision}, Verona, Italy, September 2018, pp. 32--41.

\bibitem{SceneGraphFusion}
S.-C. Wu, J.~Wald, K.~Tateno, N.~Navab, and F.~Tombari, ``Scenegraphfusion: Incremental 3d scene graph prediction from rgb-d sequences,'' \emph{Proceedings of the Conference on Computer Vision and Pattern Recognition}, pp. 7515--7525, June 2022.

\bibitem{werby2024hierarchical}
A.~Werby, C.~Huang, M.~Büchner, A.~Valada, and W.~Burgard, ``{Hierarchical Open-Vocabulary 3D Scene Graphs for Language-Grounded Robot Navigation},'' in \emph{Proceedings of the Robotics: Science and Systems}, Delft, Netherlands, July 2024.

\bibitem{wang2024llm}
J.~Wang and L.~Ke, ``Llm-seg: Bridging image segmentation and large language model reasoning,'' in \emph{Proceedings of the International Conference on Computer Vision and Pattern Recognition}, Seattle, WA, USA, June 2024, pp. 1765--1774.

\bibitem{Shi_Dao_Cai_2024}
H.~Shi, S.~D. Dao, and J.~Cai, ``Llmformer: Large language model for open-vocabulary semantic segmentation,'' \emph{International Journal of Computer Vision}, August 2024.

\bibitem{MaYZ0LZH23}
X.~Ma, S.~Yong, Z.~Zheng, Q.~Li, Y.~Liang, S.~Zhu, and S.~Huang, ``{SQA3D:} situated question answering in 3d scenes,'' in \emph{Proceedings of the International Conference on Learning Representations}, Kigali, Rwanda, May 2023.

\bibitem{AzumaMKK22}
D.~Azuma, T.~Miyanishi, S.~Kurita, and M.~Kawanabe, ``Scanqa: 3d question answering for spatial scene understanding,'' in \emph{Proceedings of the Conference on Computer Vision and Pattern Recognition}, New Orleans, LA, USA, June 2022, pp. 19\,107--19\,117.

\bibitem{AchlioptasAXEG20}
P.~Achlioptas, A.~Abdelreheem, F.~Xia, M.~Elhoseiny, and L.~J. Guibas, ``Referit3d: Neural listeners for fine-grained 3d object identification in real-world scenes,'' in \emph{Proceedings of the European Conference on Computer Vision}, Glasgow, UK, August 2020, pp. 422--440.

\bibitem{grounded3dllm}
Y.~Chen, S.~Yang, H.~Huang, T.~Wang, R.~Lyu, R.~Xu, D.~Lin, and J.~Pang, ``Grounded 3d-llm with referent tokens,'' \emph{arXiv preprint arXiv:2405.10370}, 2024.

\bibitem{3dllm}
Y.~Hong, H.~Zhen, P.~Chen, S.~Zheng, Y.~Du, Z.~Chen, and C.~Gan, ``3d-llm: Injecting the 3d world into large language models,'' in \emph{Proceedings of the Advances in Neural Information Processing Systems}, New Orleans, LA, USA, December 2023.

\bibitem{ll3da}
S.~Chen, X.~Chen, C.~Zhang, M.~Li, G.~Yu, H.~Fei, H.~Zhu, J.~Fan, and T.~Chen, ``Ll3da: Visual interactive instruction tuning for omni-3d understanding reasoning and planning,'' in \emph{Proceedings of the International Conference on Computer Vision and Pattern Recognition}, Seattle, WA, USA, June 2024, pp. 26\,428--26\,438.

\bibitem{YangC0MIFC24}
J.~Yang, X.~Chen, S.~Qian, N.~Madaan, M.~Iyengar, D.~F. Fouhey, and J.~Chai, ``{LLM-Grounder}: Open-vocabulary 3d visual grounding with large language model as an agent,'' in \emph{Proceedings of the International Conference on Robotics and Automation}, Yokohama, Japan, May 2024, pp. 7694--7701.

\bibitem{ChenMZXQYCF24}
X.~Chen, Z.~Ma, X.~Zhang, S.~Xu, S.~Qian, J.~Yang, D.~F. Fouhey, and J.~Chai, ``Multi-object hallucination in vision-language models,'' in \emph{Proceedings of the Advances in Neural Information Processing Systems}, Vancouver, BC, Canada, December 2024.

\bibitem{achiam2023gpt}
J.~Achiam, S.~Adler, S.~Agarwal, L.~Ahmad, I.~Akkaya, F.~L. Aleman, D.~Almeida, J.~Altenschmidt, S.~Altman, S.~Anadkat \emph{et~al.}, ``Gpt-4 technical report,'' \emph{arXiv preprint arXiv:2303.08774}, 2023.

\bibitem{koch2024open3dsg}
S.~Koch, N.~Vaskevicius, M.~Colosi, P.~Hermosilla, and T.~Ropinski, ``{Open3DSG}: Open-vocabulary 3d scene graphs from point clouds with queryable objects and open-set relationships,'' in \emph{Proceedings of the Conference on Computer Vision and Pattern Recognition}, Seattle, WA, USA, June 2024, pp. 14\,183--14\,193.

\bibitem{kuhn1955hungarian}
H.~W. Kuhn, ``The hungarian method for the assignment problem,'' \emph{Naval research logistics quarterly}, vol.~2, no. 1-2, pp. 83--97, 1955.

\bibitem{SceneNN}
B.~Hua, Q.~Pham, D.~T. Nguyen, M.~Tran, L.~Yu, and S.~Yeung, ``Scenenn: {A} scene meshes dataset with annotations,'' in \emph{Proceedings of the International Conference on 3D Vision}, Stanford, CA, USA, October 2016, pp. 92--101.

\bibitem{ScanNet}
A.~Dai, A.~X. Chang, M.~Savva, M.~Halber, T.~A. Funkhouser, and M.~Nie{\ss}ner, ``Scannet: Richly-annotated 3d reconstructions of indoor scenes,'' in \emph{Proceedings of the Conference on Computer Vision and Pattern Recognition}, Honolulu, HI, USA, July 2017, pp. 2432--2443.

\bibitem{ScanNetv2}
C.~Yeshwanth, Y.~Liu, M.~Nie{\ss}ner, and A.~Dai, ``Scannet++: {A} high-fidelity dataset of 3d indoor scenes,'' in \emph{Proceedings of the International Conference on Computer Vision}, Paris, France, October 2023, pp. 12--22.

\bibitem{scannet200}
D.~Rozenberszki, O.~Litany, and A.~Dai, ``Language-grounded indoor 3d semantic segmentation in the wild,'' in \emph{Proceedings of the European Conference on Computer Vision}, S.~Avidan, G.~J. Brostow, M.~Ciss{\'{e}}, G.~M. Farinella, and T.~Hassner, Eds., Tel Aviv, Israel, October 2022, pp. 125--141.

\bibitem{replica}
J.~Straub, T.~Whelan, L.~Ma, Y.~Chen, E.~Wijmans, S.~Green, J.~J. Engel, R.~Mur-Artal, C.~Ren, S.~Verma, A.~Clarkson, M.~Yan, B.~Budge, Y.~Yan, X.~Pan, J.~Yon, Y.~Zou, K.~Leon, N.~Carter, J.~Briales, T.~Gillingham, E.~Mueggler, L.~Pesqueira, M.~Savva, D.~Batra, H.~M. Strasdat, R.~D. Nardi, M.~Goesele, S.~Lovegrove, and R.~Newcombe, ``The {R}eplica dataset: A digital replica of indoor spaces,'' \emph{arXiv preprint arXiv:1906.05797}, 2019.

\bibitem{VoxbloxDiffusion}
R.~Mascaro, L.~Teixeira, and M.~Chli, ``Volumetric instance-level semantic mapping via multi-view 2d-to-3d label diffusion,'' \emph{IEEE Robotics and Automation Letters}, vol.~7, no.~2, pp. 3531--3538, 2022.

\bibitem{ModifiedVoxblox}
M.~Han, Z.~Zhang, Z.~Jiao, X.~Xie, Y.~Zhu, S.~Zhu, and H.~Liu, ``Reconstructing interactive 3d scenes by panoptic mapping and {CAD} model alignments,'' in \emph{Proceedings of the International Conference on Robotics and Automation}, Xi'an, China, May 2021, pp. 12\,199--12\,206.

\bibitem{MultiviewFusion}
L.~Wang, R.~Li, J.~Sun, X.~Liu, L.~Zhao, H.~S. Seah, C.~K. Quah, and B.~Tandianus, ``Multi-view fusion-based 3d object detection for robot indoor scene perception,'' \emph{Sensors}, vol.~19, no.~19, p. 4092, 2019.

\bibitem{IncrementalBBox}
W.~Li, J.~Gu, B.~Chen, and J.~Han, ``Incremental instance-oriented 3d semantic mapping via rgb-d cameras for unknown indoor scene,'' \emph{Discrete Dynamics in Nature and Society}, vol. 2020, pp. 1--10, 2020.

\bibitem{miao2024volumetric}
Y.~Miao, I.~Armeni, M.~Pollefeys, and D.~Barath, ``Volumetric semantically consistent 3d panoptic mapping,'' in \emph{2024 IEEE/RSJ International Conference on Intelligent Robots and Systems (IROS)}.\hskip 1em plus 0.5em minus 0.4em\relax IEEE, 2024, pp. 12\,924--12\,931.

\bibitem{campos2021orb}
C.~Campos, R.~Elvira, J.~J.~G. Rodr{\'\i}guez, J.~M. Montiel, and J.~D. Tard{\'o}s, ``Orb-slam3: An accurate open-source library for visual, visual--inertial, and multimap slam,'' \emph{IEEE Transactions on Robotics}, vol.~37, no.~6, pp. 1874--1890, 2021.

\bibitem{huang2024openins3d}
Z.~Huang, X.~Wu, X.~Chen, H.~Zhao, L.~Zhu, and J.~Lasenby, ``Openins3d: Snap and lookup for 3d open-vocabulary instance segmentation,'' in \emph{Proceedings of the European Conference on Computer Vision}, 2024, pp. 169--185.

\end{thebibliography}

\end{document}